\documentclass{bmvc2k}
\usepackage{amssymb}
\usepackage{booktabs}
\usepackage{multirow}
\usepackage{wrapfig}

\title{Restoring Without Forgetting: Continual Learning Across Image Degradations}

\addauthor{Alif Ashrafee}{aa5264@rit.edu}{1}
\addauthor{Bartosz Krawczyk}{bartosz.krawczyk@rit.edu}{1}

\addinstitution{
 Rochester Institute of Technology \\
 New York, USA
}

\runninghead{Ashrafee \& Krawczyk}{Restoring Without Forgetting}

\begin{document}

\maketitle

\begin{abstract}
Recent progress in image restoration has converged on all-in-one architectures that jointly handle multiple degradations within a single network. These methods are effective on static benchmarks but target a closed-world setting that assumes simultaneous access to every target degradation at training time. In practice, degradations are encountered sequentially as field-deployed systems progressively face new environmental conditions, and historical training data is often unavailable due to privacy or storage constraints. Accommodating a new degradation then requires either retraining on the union of all prior data, which is often costly or infeasible, or fine-tuning, which causes catastrophic forgetting. We formulate multi-degradation image restoration as a continual domain-incremental learning problem, in which degradations arrive incrementally and prior data is unavailable. Our proposed Restoring without Forgetting (RwF) framework learns a lightweight adapter for each new degradation, eliminating forgetting by construction at a fraction of the cost of dedicated per-domain networks. To isolate degradation learning from dataset variation, we construct a benchmark spanning five degradation domains under shared image content. At test time, an unsupervised routing mechanism identifies the appropriate restoration path for unknown inputs without requiring domain labels. Across the five-domain sequence, RwF improves final average PSNR over naive sequential fine-tuning by 15.25~dB and 11.83~dB on the Restormer and NAFNet backbones, respectively. The framework transfers to eleven canonical real-degradation benchmarks (3,465 images) at 89.5\% routing accuracy with only a $+0.94$~dB oracle PSNR gap, establishing, to our knowledge, the first systematic baseline for continual multi-degradation image restoration. Code, benchmark, and weights are available at: \url{https://github.com/AlifAshrafee/Restoring-Without-Forgetting}.
\end{abstract}

\section{Introduction}
\label{sec:intro}

Image restoration has advanced rapidly in recent years, propelled by attention-based and transformer architectures that exploit long-range pixel dependencies for high-fidelity reconstruction~\cite{liang2021swinir, wang2022uformer, tu2022maxim}. Networks such as Restormer~\cite{zamir2022restormer}, NAFNet~\cite{chen2022simple}, SFHformer~\cite{sfhformer}, and FFTformer~\cite{kong2023efficient} have established increasingly strong baselines across denoising, deblurring, dehazing, deraining, and low-light enhancement. Yet the dominant practice remains task-specialized: each network is trained on a single degradation type, with dedicated data, hyperparameter tuning, and model weights for every problem~\cite{lim2017enhanced, fu2017removing, cai2016dehazenet, nah2017deep, tian2020image}. This single-degradation pipeline is poorly suited to real deployment. Field-deployed sensors progressively encounter new environmental conditions over their operational lifetime, often without access to data from previously seen degradations due to storage, privacy, or licensing constraints. Naively fine-tuning an existing network on a new degradation overwrites the representations responsible for prior modes, leading to catastrophic forgetting~\cite{mccloskey1989catastrophic, french1999catastrophic}. In contrast, training a fresh network from scratch for every encountered degradation is expensive, data-hungry, and increasingly unscalable as the catalog of degradations grows. Equally absent from this paradigm is a mechanism to decide which restoration network to apply to a given input at test time. Existing pipelines implicitly assume an oracle that pairs each test image with the correct network, an assumption that breaks down for intelligent systems operating in the open world.

\begin{figure}[t]
  \centering
  \subfigure[]{%
    \includegraphics[width=0.58\textwidth]{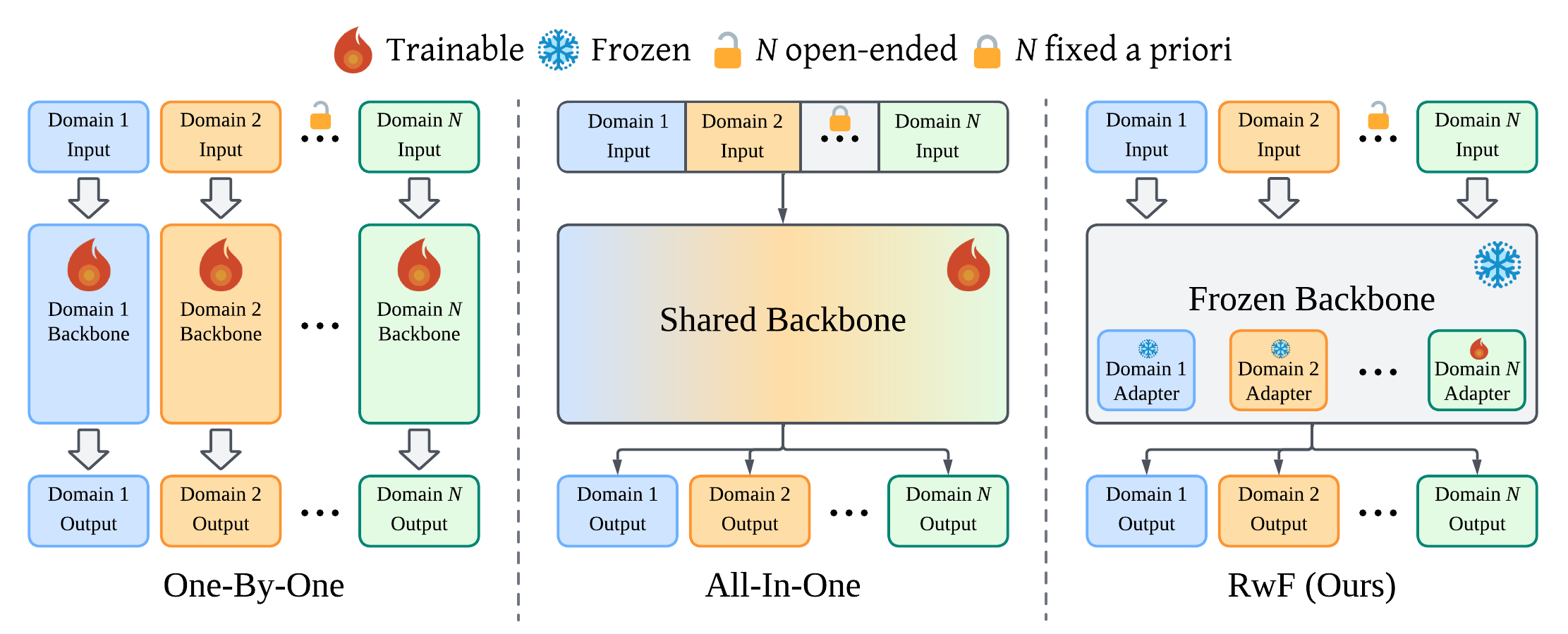}%
    \label{fig:paradigms}%
  }
  \hfill
  \subfigure[]{%
    \includegraphics[width=0.4\textwidth]{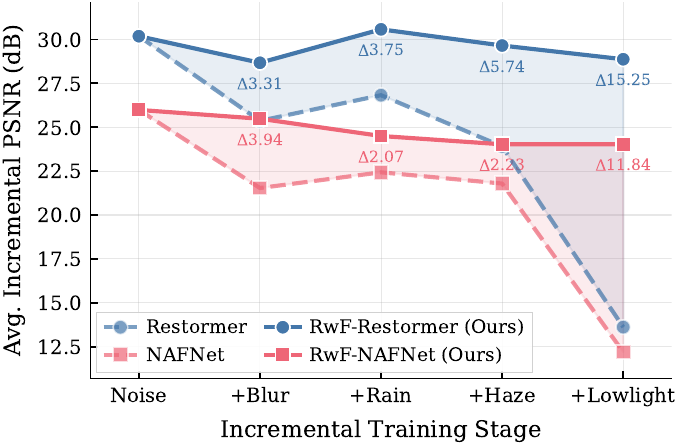}%
    \label{fig:forgetting_plot}%
  }
  \caption{\textbf{(a)} Three paradigms for multi-degradation image restoration: One-By-One (one network per degradation), All-In-One (a single jointly trained network), and RwF (Ours), which freezes a pretrained backbone and attaches lightweight degradation-specific adapters routed without supervision at inference. \textbf{(b)} Average incremental PSNR across five sequentially learned degradations (Noise, +Blur, +Rain, +Haze, +Lowlight). Sequential fine-tuning forgets as degradations accumulate, while RwF holds performance across all domains.}
  \label{fig:method_and_forgetting}
\end{figure}


To reduce this parameter redundancy, recent work consolidates multiple degradations into ``all-in-one'' networks trained jointly on the union of target tasks~\cite{li2022all, potlapalli2023promptir, zhang2023ingredient, qin2024restore, gong2025mini}. Effective as they are on closed-set benchmarks, these models presuppose simultaneous access to every target degradation in a single joint training run, with the degradation set fixed at design time. Extending one to a newly observed degradation then requires reassembling the full union of past and present data and retraining, which is costly at scale and infeasible when prior data cannot be retained. Weather-specialized restoration~\cite{valanarasu2022transweather} narrows the problem to atmospheric phenomena and offers no clear extension to the sensor-side artifacts (noise, low-light, blur) that deployed perception stacks routinely encounter.


The challenge of accumulating capability over a sequential data stream while preserving previously learned competence is the central problem of continual learning (CL)~\cite{parisi2019continual, wang2024comprehensive}. Domain-incremental learning (DIL)~\cite{van2019three} in particular targets the setting where new domains arrive sequentially and domain identity is unavailable at test time. Yet CL research remains predominantly evaluated in discriminative, class-incremental settings on classification benchmarks, with limited engagement with dense, pixel-level generative tasks. Image restoration, acutely sensitive to distributional shifts across degradation types and demanding faithful per-pixel reconstruction, is conspicuously underexplored under the CL framework.

The central challenge is therefore not only to restore images from diverse degradations, but to acquire these capabilities sequentially, without retaining past data and without knowing the degradation label at test time. Motivated by this, we propose RwF (Restoring without Forgetting), a parameter-efficient framework for continual multi-degradation restoration. RwF freezes a pretrained restoration backbone that provides a strong natural-image prior and incrementally attaches a lightweight low-rank adapter for each new degradation, isolating degradation-specific parameters so that previously learned degradations are preserved exactly. To resolve degradation identity at inference, a prototype-matching mechanism routes each input to the appropriate adapter without domain labels. As shown in Fig.~\ref{fig:method_and_forgetting}(b), naive sequential fine-tuning (FT) of strong backbones suffers severe forgetting as degradations accumulate, whereas RwF preserves performance across the full sequence, improving final average PSNR by up to 15.25~dB. Our main contributions are as follows:

\begin{itemize}
    \item \textbf{A continual learning formulation of multi-degradation image restoration.} We frame multi-degradation restoration as a domain-incremental learning problem in which qualitatively distinct degradations arrive sequentially and degradation identity is unknown at test time, a setting that has remained largely unexplored within the broader continual learning literature.

    \item \textbf{A parameter-efficient framework with unsupervised inference-time routing.} We introduce a backbone-frozen architecture that eliminates forgetting by construction through degradation-specific low-rank adapters at a fraction of the cost of dedicated per-domain networks, and routes test inputs to the appropriate restoration path without domain labels via a lightweight prototype-matching mechanism.

    \item \textbf{A content-controlled benchmark and sim-to-real evaluation.} We construct a five-domain benchmark spanning blur, noise, haze, rain, and low-light over a shared image corpus, isolating degradation learning from dataset variation, and complement it with evaluations on canonical real-degradation benchmarks to demonstrate sim-to-real transferability of the learned adapters.
\end{itemize}

\section{Related Work}
\label{sec:related_work}

\subsection{Image Restoration}

Modern image restoration is dominated by transformer-based architectures that capture long-range dependencies for high-fidelity reconstruction. SwinIR~\cite{liang2021swinir} adapted shifted-window self-attention to restoration, Restormer~\cite{zamir2022restormer} reorganized attention along the channel dimension for efficient high-resolution processing, NAFNet~\cite{chen2022simple} demonstrated that careful simplification of nonlinear activations yields a competitive convolutional baseline, FFTformer~\cite{kong2023efficient} exploited frequency-domain representations, and SFHformer~\cite{sfhformer} fused spatial and frequency cues within a unified architecture. Despite their strong per-task performance, these networks remain single-degradation specialists, requiring a distinct trained instance for every problem.

To consolidate this redundancy, the all-in-one paradigm trains a single network jointly on the union of target degradations. AirNet~\cite{li2022all} and TransWeather~\cite{valanarasu2022transweather} pair a shared backbone with degradation-aware modules, PromptIR~\cite{potlapalli2023promptir} learns visual prompts that adapt the network to the input degradation, IDR~\cite{zhang2023ingredient} decomposes degradations into shared ingredients, AdaIR~\cite{cui2025adair} modulates frequency-domain features adaptively, and RAM~\cite{qin2024restore} brings masked image modeling to blind multi-degradation restoration. These methods assume joint access to every target degradation at training time, with the full degradation set specified in advance. Subsequent work scales this paradigm without departing from the assumption. FoundIR~\cite{li2025foundir} trains on a million-image corpus through an incremental schedule that serves optimization stability rather than continual acquisition. DegAE~\cite{liu2023degae} synthesizes degradations to pretrain transferable low-level features for downstream fine-tuning. DCPT~\cite{hu2025universal} pretrains a degradation classifier over a predetermined set to initialize a joint restoration model. A nascent line of work has begun to examine continual learning in restoration more directly. CauSiam~\cite{cui2025continual} performs continual test-time adaptation for defocus deblurring but not offline domain-incremental training. MINI~\cite{gong2025mini} adds new restoration capabilities through a meta-convolution module, though its capacity is bounded by an embedding pool that must be fixed at design time and cannot be expanded as degradations accumulate. None of these works report a forgetting protocol across heterogeneous degradation types under unsupervised inference. We position RwF in this gap, as a domain-incremental framework that retains no past data, assigns each degradation its own restoration path under strict parameter isolation, and selects among paths by label-free routing.

\subsection{Domain-Incremental Learning}

Domain-incremental learning addresses sequential adaptation across new data domains without access to domain identity at inference~\cite{van2019three, wang2024comprehensive}. The dominant strategies fall into three families: regularization of important parameters~\cite{kirkpatrick2017overcoming, zenke2017continual}, replay of stored exemplars~\cite{chaudhry2018efficient, chaudhry2019tiny, jeeveswaran2024gradual}, and parameter isolation~\cite{li2017learning, mallya2018packnet}. Replay carries storage and privacy costs that are impractical at high resolution~\cite{xu2025fr2seg, wang2023rethinking}, motivating a recent shift toward exemplar-free parameter-isolation approaches over frozen pretrained backbones~\cite{mcdonnell2023ranpac, zhou2024expandable, zhou2025revisiting, sun2025mos}. AdaptFormer~\cite{chen2022adaptformer} established adapters as a parameter-efficient mechanism for transformer fine-tuning, and methods such as SOYO~\cite{wang2025boosting} and DUCT~\cite{zhou2025dual} have refined exemplar-free domain selection and embedding-space calibration. Beyond classification, DIL has been extended to dense semantic segmentation~\cite{rui2023dilrs}, but the literature remains overwhelmingly discriminative. Pixel-level generative tasks such as image restoration are largely absent, which is the gap our work addresses.

\section{Methodology}
\label{sec:method}

\subsection{Problem Formulation}
\label{subsec:formulation}

We cast multi-degradation image restoration as a domain-incremental learning (DIL) problem~\cite{van2019three}. Throughout, \emph{degradation} refers to a physical corruption operator (motion blur, sensor noise, atmospheric haze, and so on), and \emph{domain} refers to the distribution of images it produces. Thus, each degradation type constitutes a distinct domain, and the domains are revealed to the model one at a time. Let $x \in \mathbb{R}^{H \times W \times 3}$ denote a clean image drawn from a content distribution $p(x)$, and let a degradation be described by a (possibly stochastic) operator $\mathcal{T}: \mathbb{R}^{H \times W \times 3} \rightarrow \mathbb{R}^{H \times W \times 3}$ that maps a clean image to a degraded observation $y = \mathcal{T}(x)$. A restoration model seeks an approximate inverse that recovers $x$ from $y$. We consider a sequence of $T$ degradation domains $\mathcal{D}_1, \mathcal{D}_2, \ldots, \mathcal{D}_T$, each induced by a distinct operator $\mathcal{T}_t$ and instantiated as a set of paired samples $\mathcal{D}_t = \{(x_i, y_i^{(t)})\}_{i=1}^{N_t}$ with $y_i^{(t)} = \mathcal{T}_t(x_i)$. The operators span qualitatively different physical processes. In this work we instantiate five canonical degradations through the following forward models:
\begin{align}
    \text{Noise:}    \quad & \mathcal{T}_{\mathrm{n}}(x) = x + \mathbf{n}, \quad \mathbf{n} \sim \mathcal{N}(0, \sigma^2 \mathbf{I}); \label{eq:noise}\\
    \text{Motion blur:} \quad & \mathcal{T}_{\mathrm{b}}(x) = \mathbf{k}_{L,\theta,C} \otimes x + \mathbf{n}_b; \label{eq:blur}\\
    \text{Haze:}     \quad & \mathcal{T}_{\mathrm{h}}(x) = x \odot \mathbf{t} + A\,(1 - \mathbf{t}), \quad \mathbf{t} = e^{-\beta \mathbf{d}}; \label{eq:haze}\\
    \text{Rain:}     \quad & \mathcal{T}_{\mathrm{r}}(x) = (1 - v)\big(x + \alpha \textstyle\sum_{l=1}^{L} \mathbf{s}_l\big) + v\,A_r; \label{eq:rain}\\
    \text{Low-light:}\quad & \mathcal{T}_{\mathrm{l}}(x) = \tfrac{1}{k}\,\mathrm{Pois}\big(k\,x^{\gamma}\big) + \mathbf{n}_r, \label{eq:lowlight}
\end{align}
where $\otimes$ denotes convolution with a motion blur kernel $\mathbf{k}_{L,\theta,C}$ rasterized from a parametric trajectory of length $L$, direction $\theta$, and curvature $C$; $\odot$ is the Hadamard product; $\mathbf{t}$ is the transmission map governed by per-pixel scene depth $\mathbf{d}$ and scattering coefficient $\beta$; $A$ is the global atmospheric light; $\mathbf{s}_l$ are per-layer rain streak maps at increasing depth planes (with $\alpha$ controlling streak intensity, $v$ the atmospheric veiling, and $A_r$ the gray atmospheric light from droplet scattering); $\gamma > 1$ controls illumination attenuation; $\mathrm{Pois}(\cdot)$ denotes a pixelwise Poisson resampling with photon gain $k$; and $\mathbf{n}_b$, $\mathbf{n}_r$ are Gaussian noise terms modeling post-blur sensor readout and read noise, respectively. These models follow established degradation formulations in the restoration literature (Sec.~\ref{subsec:synthesis}).

In the DIL setting, domains are encountered strictly sequentially: at stage $t$ only $\mathcal{D}_t$ is accessible, while all prior domains $\mathcal{D}_{<t}$ are unavailable and no exemplars are retained. Let $f_\theta$ denote a restoration network with parameters $\theta$. At the initial stage the backbone is trained on $\mathcal{D}_1$ under an $\ell_1$ reconstruction loss, as is standard for restoration~\cite{zamir2022restormer, chen2022simple}. For any later domain, unconstrained fine-tuning from the previous parameters minimizes the current-domain loss but overwrites the representations supporting earlier domains, the hallmark of catastrophic forgetting~\cite{mccloskey1989catastrophic, french1999catastrophic}. After traversing all $T$ domains, the model parameters $\Theta_T$ should minimize the average reconstruction loss over every domain seen so far,
\begin{equation}
    \mathcal{L}_{\mathrm{CL}} = \frac{1}{T}\sum_{t=1}^{T} \frac{1}{N_t}\sum_{i=1}^{N_t} \big\| f_{\Theta_T}(y_i^{(t)}) - x_i \big\|_1,
    \label{eq:cl_objective}
\end{equation}
subject to the constraint that domain data is never jointly available, which separates our setting from all-in-one joint training~\cite{li2022all, potlapalli2023promptir, zhang2023ingredient}, and that no past samples are stored, which separates it from replay-based continual learning~\cite{chaudhry2018efficient, chaudhry2019tiny}. Crucially, the domain index $t$ is also withheld at test time, so the model must additionally infer which restoration behavior to apply for each input. We resolve the training-time tension in Sec.~\ref{subsec:framework} and the test-time routing in Sec.~\ref{subsec:routing}.

\subsection{Pretrained Backbone and Degradation-Isolated Learning}
\label{subsec:backbone}

Modern continual learning increasingly builds on a strong pretrained backbone used as a frozen, general-purpose feature extractor, with adaptation confined to lightweight modules~\cite{mcdonnell2023ranpac, zhou2024expandable, sun2025mos}. Where discriminative CL takes an ImageNet~\cite{russakovsky2015imagenet}-pretrained encoder as this prior, we seek its restoration analogue, and image denoising is a natural candidate: a denoiser must learn the statistics of clean natural images to separate signal from corruption, which is precisely the prior that plug-and-play and regularization-by-denoising methods reuse as a universal proximal operator across inverse problems~\cite{venkatakrishnan2013plug, zhang2021plug}. A denoising-optimized backbone therefore captures degradation-agnostic restoration primitives that transfer to other degradations. We accordingly take the publicly released Gaussian color denoising weights of Restormer~\cite{zamir2022restormer} off the shelf as our backbone $f_{\theta_1^*}$ and freeze them throughout, so they serve as the shared, fixed substrate over which all subsequent degradations are learned. The same scheme transfers without modification to NAFNet~\cite{chen2022simple} using its released denoising weights.

The DIL formulation idealizes a domain as differing from another only in its degradation operator, but real benchmarks violate this: each is collected for a single degradation and carries its own clean-content distribution (street scenes, indoor scenes, hazy landscapes). Training one restoration path per real dataset therefore conflates the degradation operator with the dataset content, and an adapter cannot tell which it is correcting for. To make this precise, let the clean content be a latent $c \sim p(c)$ and write $y = \mathcal{T}_t(c)$, with the frozen backbone producing a routing embedding $e = g_{\theta_1^*}(y)$. Under a multi-dataset construction, domain $t$ draws content from its own distribution $p_t(c)$, so the degraded marginal $q_t(y) = \int p(y \mid c; \mathcal{T}_t)\, p_t(c)\, dc$ varies with $t$ through both $\mathcal{T}_t$ and $p_t(c)$. In the counterfactual where all operators are identical, $\mathcal{T}_t \equiv \mathcal{T}$, any remaining dependence of $e$ on $t$ is purely content-driven, and the mutual information $I(e; t)$ decomposes as
\begin{equation}
    I(e; t) \;=\; \underbrace{I_{\mathrm{deg}}(e; t)}_{\text{from } \{\mathcal{T}_t\}} \;+\; \underbrace{I_{\mathrm{content}}(e; t)}_{\text{from } \{p_t(c)\}}, \qquad I_{\mathrm{content}}(e; t) > 0 \;\;\text{whenever}\;\; p_t(c) \not\equiv p_{t'}(c).
    \label{eq:mi_decomp}
\end{equation}
The router can then reach high domain-identification accuracy by recognizing content rather than degradation, and each path absorbs the dataset-specific prior $p_t(c)$ into its learned correction, entangling it with the degradation inverse $\mathcal{T}_t^{-1}$. The remedy is to draw all domains from one shared corpus, $p_t(c) = p(c)$ for all $t$, defining each domain solely by its operator,
\begin{equation}
    \mathcal{D}_t = \big\{ (x_i,\, y_i^{(t)}) : x_i \sim p(x),\; y_i^{(t)} = \mathcal{T}_t(x_i) \big\}.
    \label{eq:shared_corpus}
\end{equation}
Now the counterfactual $\mathcal{T}_t \equiv \mathcal{T}$ renders $e$ identically distributed across domains, forcing $I_{\mathrm{content}}(e; t) = 0$ and hence
\begin{equation}
    I(e; t) \;=\; I_{\mathrm{deg}}(e; t).
    \label{eq:isolated}
\end{equation}
All domain-distinguishing information, and therefore all routing accuracy, is then attributable to the degradation operators alone. Symmetrically, since the clean targets share a common manifold, each path is optimized to invert its operator $\mathcal{T}_t$ without absorbing a content prior. This identifiability is why our benchmark is synthesized over a shared corpus rather than assembled from heterogeneous real datasets.

\subsection{Degradation Synthesis}
\label{subsec:synthesis}

To instantiate Eq.~\eqref{eq:shared_corpus}, we apply the forward models of Eqs.~\eqref{eq:noise}--\eqref{eq:lowlight} to a single clean corpus (DIV2K~\cite{agustsson2017ntire}), producing one paired dataset per degradation in which every domain shares identical clean source images. Each operator is parameterized to match the degradation characteristics reported in the corresponding restoration literature, so that synthetic samples approximate their real-world counterparts in appearance and difficulty. Additive Gaussian noise follows the AWGN model used throughout learned denoising~\cite{zhang2017beyond}.
Motion blur kernels are rasterized from parametric trajectories of explicit length, direction, and curvature following~\cite{boracchi2012modeling}, with mild post-blur Gaussian sensor noise to mimic long-exposure handheld capture. Haze is rendered via the Koschmieder atmospheric scattering model with per-pixel transmission derived from monocular depth estimated by MiDaS~\cite{ranftl2020towards}, matching the synthesis convention of standard dehazing benchmarks~\cite{li2018benchmarking}. Rain is synthesized as a multi-layer composite in which each layer represents a depth plane with progressively shorter, thinner, and more strongly out-of-focus streaks, followed by atmospheric veiling from suspended droplet scattering, combining~\cite{garg2007vision} with the multi-layer rendering of~\cite{yang2017deep}. Low-light is generated by gamma-based illumination attenuation followed by signal-dependent Poisson shot noise and signal-independent Gaussian read noise, following the See-in-the-Dark sensor model~\cite{chen2018learning} and consistent with low-light enhancement datasets~\cite{wei2018deep}.

\subsection{Restoration Paths via Low-Rank Adaptation}
\label{subsec:framework}

Given the frozen backbone $f_{\theta_1^*}$, we learn each new degradation by attaching compact, trainable modules while leaving the backbone untouched. The atomic module is a low-rank adapter that applies a bottleneck residual correction to a block's features,
\begin{equation}
    \mathcal{A}_{\phi}(\mathbf{z}) = \mathbf{z} + s\,\mathbf{W}_{\mathrm{up}}\, \sigma\!\big(\mathbf{W}_{\mathrm{down}}\, \mathrm{LN}(\mathbf{z})\big),
    \label{eq:adapter}
\end{equation}
where $\mathbf{z} \in \mathbb{R}^{d}$ is a feature vector, $\mathbf{W}_{\mathrm{down}} \in \mathbb{R}^{r \times d}$ and $\mathbf{W}_{\mathrm{up}} \in \mathbb{R}^{d \times r}$ form a rank-$r$ bottleneck with $r \ll d$, $\sigma$ is a ReLU, $\mathrm{LN}$ is layer normalization, and $s$ is a fixed scale. Both projections are linear and act pointwise across the channel dimension. The trainable parameters are $\phi = \{\mathbf{W}_{\mathrm{down}}, \mathbf{W}_{\mathrm{up}}\}$, totaling $2dr + d$ per module. Following LoRA-style initialization~\cite{hu2022lora, he2015delving}, $\mathbf{W}_{\mathrm{down}}$ is Kaiming-initialized and $\mathbf{W}_{\mathrm{up}}$ is set to zero, so each adapter is the identity at insertion and leaves the pretrained features unperturbed until training begins.

A single adapter only corrects one block, whereas inverting a degradation requires coordinated corrections along the entire encode--decode trajectory. We therefore instantiate an adapter at every block of the backbone and define a restoration path as the complete, network-spanning set of these per-block adapters for a given degradation,
\begin{equation}
    \Phi_t = \{\phi_{t,1}, \phi_{t,2}, \ldots, \phi_{t,L}\},
    \label{eq:restoration_path}
\end{equation}
where $L$ is the number of blocks. For a Transformer block, the path-augmented forward pass wraps the post-FFN residual,
\begin{align}
    \mathbf{z}_l' &= \mathbf{z}_l + \mathrm{Attn}(\mathrm{LN}(\mathbf{z}_l)), &
    \mathbf{z}_{l+1} &= \mathcal{A}_{\phi_{t,l}}\!\big(\mathbf{z}_l' + \mathrm{FFN}(\mathrm{LN}(\mathbf{z}_l'))\big),
    \label{eq:adapted_forward}
\end{align}
with $\mathrm{Attn}$ and $\mathrm{FFN}$ frozen. For convolutional blocks, the same injection follows the channel-mixing stage. Because a path is defined purely at the block level through the channel dimension $d_l$, the construction is backbone-agnostic, and we validate it on both a Transformer (Restormer~\cite{zamir2022restormer}) and a fully convolutional network (NAFNet~\cite{chen2022simple}) under identical path definitions. The full restoration network for domain $t$ is $f_{\theta_1^*, \Phi_t}$: a shared frozen backbone routed through one degradation-specific path.

Sequential learning enforces strict parameter isolation. At domain $t$, the backbone and all committed paths $\Phi_{<t}$ are frozen and only a freshly initialized $\Phi_t$ is optimized,
\begin{equation}
    \Phi_t^* = \arg\min_{\Phi_t} \frac{1}{N_t}\sum_{i=1}^{N_t} \big\| f_{\theta_1^*, \Phi_t}(y_i^{(t)}) - x_i \big\|_1.
    \label{eq:path_training}
\end{equation}
Upon convergence, $\Phi_t^*$ is committed to a frozen bank and the procedure repeats. The final model state $\Theta_T = \{\theta_1^*\} \cup \{\Phi_1^*, \ldots, \Phi_T^*\}$ thus occupies mutually disjoint parameter subspaces for distinct domains, so the cross-domain gradient vanishes and forgetting is exactly zero by construction,
\begin{equation}
    \frac{\partial \Phi_t^*}{\partial \Phi_j^*} = \mathbf{0}\quad (j \neq t) \;\;\Longrightarrow\;\; \mathcal{F}_{j,t} = 0\quad (j < t),
    \label{eq:zero_forgetting}
\end{equation}
achieved without regularization penalties or replay, at a per-domain overhead of only $|\Phi_t| \approx 0.03\,|\theta_1^*|$ that scales linearly in the number of domains rather than duplicating the backbone. Because each path $\Phi_t$ and its prototype are estimated independently against the frozen backbone, with no gradient coupling across domains, the model state and the prototype bank are identical under any permutation of the domain order.

\begin{figure*}[t]
  \centering
  \includegraphics[width=\textwidth]{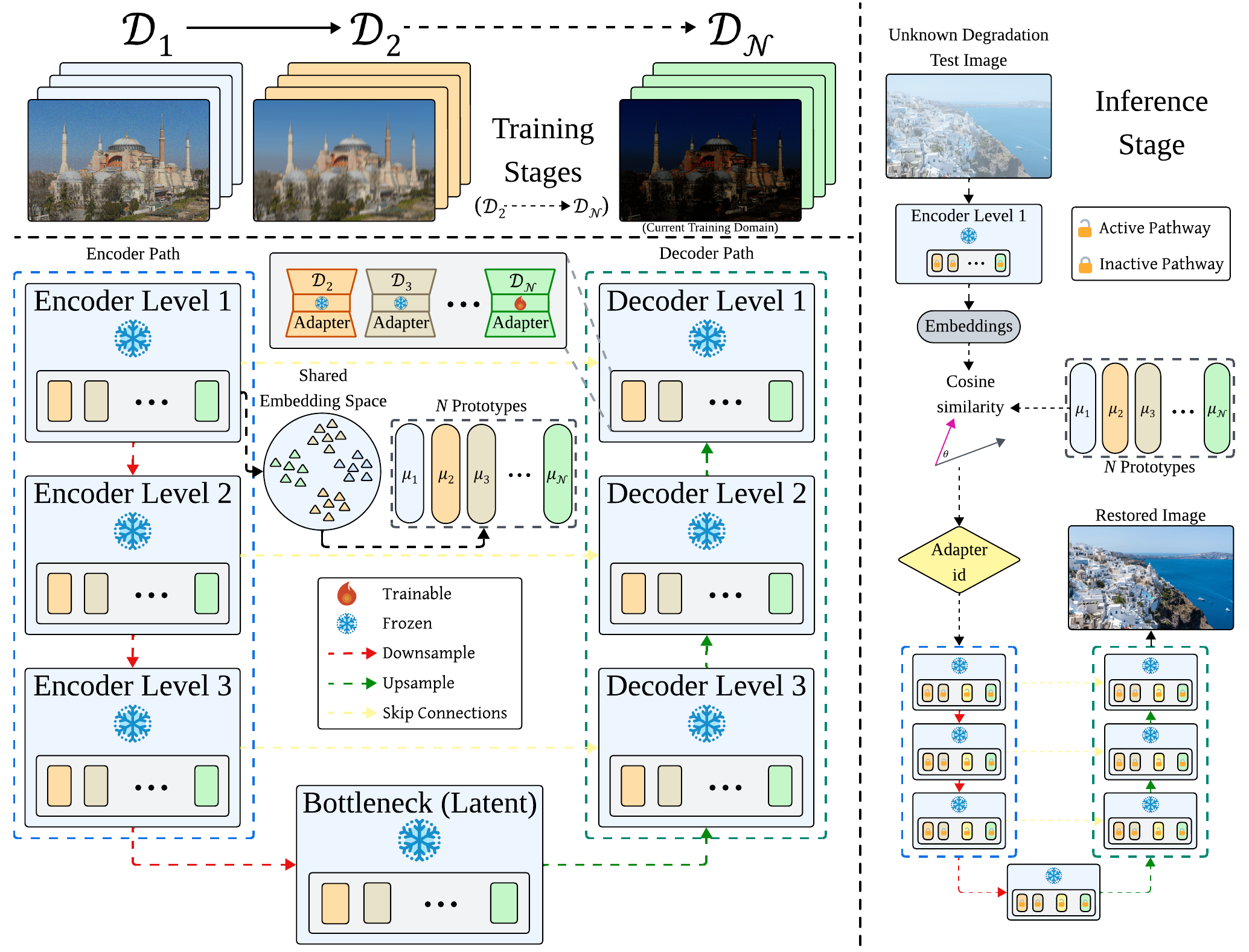}
  \caption{\textbf{RwF Framework:} a denoising-pretrained encoder--decoder backbone is frozen, and each new degradation trains a network-spanning path of low-rank adapters under strict parameter isolation, leaving prior paths and the backbone unchanged. A domain prototype is formed by pooling first-encoder-stage features. At inference, an unknown input is routed by cosine similarity to the matching path, which restores the image.}
  \label{fig:framework}
\end{figure*}

\subsection{Prototype-Based Degradation Routing}
\label{subsec:routing}

Because the domain index is withheld at test time, the model must decide which restoration path to engage for each input. We route without any auxiliary classifier or extra trainable parameters by exploiting the geometry of the frozen backbone's feature space. A central design question is where along the backbone to read out features for prototyping. In classification CL, the deepest pre-classifier features are canonical, on the intuition that representational abstraction sharpens with depth. We find this prescription inverts for cross-degradation DIL: a restoration backbone is optimized to suppress the degradation signal en route to a clean reconstruction, so each successive encoder block, by design, attenuates precisely the cues a router needs. By the bottleneck stage, distinct degradations have largely converged toward a shared clean-image attractor and become difficult to discriminate, whereas early-encoder features still carry visible imprints of the operator. We empirically corroborate this depth-discrimination trade-off in Sec.~\ref{sec:experiments} (Tab.~\ref{tab:ablation_depth}) and accordingly read out at the output of the last block of the first encoder stage, denoted $g^{(1)}_{\theta_1^*}(\cdot)$.

A second design choice is how to aggregate this spatial feature map into a fixed-length embedding. Different degradations leave signatures in different orders of spatial statistics. Noise inflates per-channel standard deviation without altering channel means, haze flattens both, rain streaks introduce structured high spatial variance over an otherwise clean background, and low-light shifts mean intensities. A first-order summary (Global Average Pooling, GAP) alone discards the variance signal, while a joint normalization of mean and standard deviation lets the larger-magnitude component dominate the cosine similarity. We therefore compute both moments and normalize them independently. Let $\mathbf{F} = g^{(1)}_{\theta_1^*}(y) \in \mathbb{R}^{C \times H \times W}$, and define
\begin{equation}
    \boldsymbol{\mu}_F = \frac{1}{HW}\sum_{h,w}\mathbf{F}_{:,h,w}, \qquad \boldsymbol{\sigma}_F = \sqrt{\frac{1}{HW}\sum_{h,w}\big(\mathbf{F}_{:,h,w} - \boldsymbol{\mu}_F\big)^2}.
    \label{eq:gap_std}
\end{equation}
We $\ell_2$-normalize each moment independently and concatenate them into a single embedding,
\begin{equation}
    \mathbf{e} = \Big[\,\boldsymbol{\mu}_F / \|\boldsymbol{\mu}_F\|_2 \;\big\|\; \boldsymbol{\sigma}_F / \|\boldsymbol{\sigma}_F\|_2\,\Big] \in \mathbb{R}^{2C}.
    \label{eq:embedding}
\end{equation}
The domain prototype is the mean embedding over a subset $\mathcal{S}_t$ of the domain's training samples, $\boldsymbol{\mu}_t = \tfrac{1}{|\mathcal{S}_t|}\sum_{i \in \mathcal{S}_t}\mathbf{e}_i$. All embeddings are extracted through the frozen backbone with no path engaged, so prototypes inhabit a single shared space and cross-domain cosine comparisons are well-posed. At inference, an unknown input $y^*$ is embedded via Eqs.~\eqref{eq:gap_std}--\eqref{eq:embedding} and routed to the path whose prototype it most resembles,
\begin{equation}
    t^* = \arg\max_{t \in \{1, \ldots, T\}} \frac{\langle \mathbf{e}^*, \boldsymbol{\mu}_t \rangle}{\|\mathbf{e}^*\|_2 \,\|\boldsymbol{\mu}_t\|_2},
    \label{eq:routing}
\end{equation}
after which the image is restored by a forward pass through the selected path, $\hat{x}^* = f_{\theta_1^*, \Phi_{t^*}}(y^*)$. In practice we compute each prototype from $\rho = 50\%$ of the domain's training samples, halving the prototype-extraction cost with negligible impact on routing accuracy. Fig.~\ref{fig:framework} summarizes the complete training and inference pipeline.

\subsection{Continual Learning Evaluation Protocol}
\label{subsec:evaluation}

Standard restoration benchmarks score a model on one degradation, or under joint access to all degradations, and thus cannot reveal how performance on earlier degradations evolves as later ones are learned. We adopt the continual learning evaluation framework of~\cite{chaudhry2018efficient}, organized around an evaluation matrix $\mathbf{R} \in \mathbb{R}^{T \times T}$ whose entry $R_{b,j}$ is the restoration quality (PSNR or SSIM) on domain $\mathcal{D}_j$ after the model has been trained through domain $\mathcal{D}_b$, defined for $j \le b$. From this lower-triangular matrix we report three summaries,
\begin{equation}
    A_B = \frac{1}{T}\sum_{j=1}^{T} R_{T,j}, \qquad
    \bar{A} = \frac{1}{T}\sum_{b=1}^{T}\frac{1}{b}\sum_{j=1}^{b} R_{b,j}, \qquad
    \mathcal{F} = \frac{1}{T-1}\sum_{j=1}^{T-1}\big(R_{j,j} - R_{T,j}\big),
    \label{eq:cl_metrics}
\end{equation}
where $A_B$ is the average final quality across all domains after the last stage, $\bar{A}$ is the average incremental quality over all stages and thus reflects stability throughout the sequence, and $\mathcal{F}$ is the average forgetting, the mean drop on each domain relative to its peak, with $\mathcal{F} = 0$ denoting perfect retention.

\section{Experiments}
\label{sec:experiments}

\subsection{Experimental Setup}
\label{subsec:setup}

\paragraph{Backbones and adapter paths.} We instantiate RwF on two restoration backbones spanning the dominant architectural families: a Transformer, Restormer~\cite{zamir2022restormer}, and a fully convolutional network, NAFNet~\cite{chen2022simple}. For each, we initialize from publicly released Gaussian-denoising weights (Restormer's blind Gaussian color denoising checkpoint; NAFNet's SIDD width-32 checkpoint) and freeze the backbone throughout. These choices instantiate $\mathcal{D}_1$ directly: the pretrained backbone serves as the domain-1 restorer (Sec.~\ref{subsec:backbone}), and only adapter paths are trained for $\mathcal{D}_2$--$\mathcal{D}_5$. All adapter slots use a bottleneck dimension of $r{=}64$, input LayerNorm, dropout $0.1$, and a fixed scale $s{=}1.0$, applied within the FFN of each Restormer block and the channel-mixing branch of each NAFNet block. Per-domain adapter parameters total $0.98$M for Restormer ($3.63\%$ of the $27.11$M backbone) and $1.29$M for NAFNet ($4.25\%$ of the $30.45$M backbone), scaling linearly with $T$ rather than duplicating the backbone. Aggregated over the full five-domain sequence, RwF-Restormer totals one shared $27.11$M backbone plus four adapters ($\mathcal{D}_2$--$\mathcal{D}_5$), roughly $31$M parameters, versus $5{\times}27.11{=}135.6$M for five separate per-degradation specialists (over $4{\times}$ fewer), while routing adds only a single stage-1 forward pass at inference and no trainable parameters. Restormer's bias-free denoising LayerNorm~\cite{mohan2019robust} caused exploding gradients under the larger degradation shift, so we reintroduced the bias term. NAFNet required no analogous adjustment.

\paragraph{Training and evaluation protocol.} Each domain stage is trained for $100$k iterations with $\ell_1$ pixel loss, AdamW (lr $2{\times}10^{-4}$, weight decay $10^{-4}$, $\beta{=}(0.9,0.999)$), and a single-cycle cosine schedule decaying to $10^{-6}$, on $224{\times}224$ patches with batch size $8$ and standard geometric augmentations. The starting learning rate is reduced below the from-scratch defaults of Restormer ($3{\times}10^{-4}$) and NAFNet ($10^{-3}$) because we fine-tune over a pretrained backbone. Identical hyperparameters are used for the sequential FT baseline and the RwF variant of each backbone; the only difference is which parameters are unfrozen. We deliberately omit progressive patch-size training, larger architectural variants, and extended schedules common in single-task restoration recipes, since our aim is to establish a continual learning baseline rather than a state-of-the-art per-task result. Each run uses a single NVIDIA A100 80GB GPU. We evaluate under two settings. The \emph{oracle} setting supplies the ground-truth domain identity at test time, and the \emph{domain-agnostic} setting reflects the true DIL scenario, where the domain label is withheld and the prototype router (Sec.~\ref{subsec:routing}) selects an adapter from the input alone. We report PSNR~\cite{hore2010image} and SSIM~\cite{wang2004image}, summarized by the metrics $A_B$, $\bar{A}$, and $\mathcal{F}$ (Sec.~\ref{subsec:evaluation}). Following denoising-evaluation convention, $\mathcal{D}_1$ is evaluated on CBSD68~\cite{martin2001database} (the canonical test corpus for Restormer's released denoising weights), with $\mathcal{D}_1$ prototypes extracted on DIV2K-synthesized noise images to preserve a common content distribution across prototypes (Eq.~\eqref{eq:isolated}). $\mathcal{D}_2$--$\mathcal{D}_5$ are evaluated on held-out DIV2K-synthesized test sets.

\paragraph{Baselines.} We compare RwF against four references. Sequential FT of the same two backbones provides the natural reference for catastrophic forgetting under unconstrained parameter updates. We add two regularization-based continual learning baselines, EWC~\cite{kirkpatrick2017overcoming} and LwF~\cite{li2017learning}, applied to both backbones under conditions identical to sequential FT, with their regularization strengths tuned once at the $\mathcal{D}_1{\to}\mathcal{D}_2$ transition and held fixed thereafter. As these methods share a single model, we evaluate them against RwF in the oracle setting. We further report a Joint all-in-one model trained on the union of all five domains as a non-continual reference. We do not retrain specialized all-in-one networks (PromptIR~\cite{potlapalli2023promptir}, AirNet~\cite{li2022all}, IDR~\cite{zhang2023ingredient}, AdaIR~\cite{cui2025adair}, TransWeather~\cite{valanarasu2022transweather}, DCPT~\cite{hu2025universal}) under our sequential protocol, treating them as complementary closed-world methods rather than direct competitors. Deployment on canonical real-degradation benchmarks is evaluated in Sec.~\ref{subsec:agnostic}.

\subsection{Comparison with Continual Learning Baselines}
\label{subsec:main_comparison}

\begin{table}[ht]
  \centering
  \caption{Continual learning comparison across the five-domain sequence ($\mathcal{D}_1{\rightarrow}\mathcal{D}_5$: noise, blur, rain, haze, low-light) under the \emph{oracle} setting.}
  \label{tab:main_results}
  \begin{tabular}{l l ccc ccc}
    \toprule
    & & \multicolumn{3}{c}{PSNR (dB)} & \multicolumn{3}{c}{SSIM} \\
    \cmidrule(lr){3-5} \cmidrule(lr){6-8}
    Method & Backbone
    & $A_B\!\uparrow$ & $\bar{A}\!\uparrow$ & $\mathcal{F}\!\downarrow$
    & $A_B\!\uparrow$ & $\bar{A}\!\uparrow$ & $\mathcal{F}\!\downarrow$ \\
    \midrule
    \multirow{2}{*}{Sequential FT}
      & NAFNet    & 12.20 & 20.79 & 15.83 & 0.5697 & 0.6262 & 0.3287 \\
      & Restormer & 13.62 & 24.16 & 22.05 & 0.6148 & 0.7102 & 0.3705 \\
    \midrule
    \multirow{2}{*}{EWC}
      & NAFNet    & 11.63 & 20.78 & 16.00 & 0.5376 & 0.6256 & 0.3366 \\
      & Restormer & 12.00 & 21.26 & 14.43 & 0.4521 & 0.6133 & 0.3193 \\
    \midrule
    \multirow{2}{*}{LwF}
      & NAFNet    & 18.10 & 22.14 & 0.22  & 0.5559 & 0.6591 & 0.0305 \\
      & Restormer & 17.59 & 23.25 & 3.04  & 0.5361 & 0.6922 & 0.1351 \\
    \midrule
    \multirow{2}{*}{Joint (All-in-one)}
      & NAFNet    & 25.49 & -- & -- & 0.7809 & -- & -- \\
      & Restormer & 26.72 & -- & -- & 0.8230 & -- & -- \\
    \midrule
    \multirow{2}{*}{\textbf{RwF (Ours)}}
      & NAFNet    & 24.03 & 24.80 & \textbf{0.00} & 0.8003 & 0.7722 & \textbf{0.0000} \\
      & Restormer & \textbf{28.87} & \textbf{29.59} & \textbf{0.00} & \textbf{0.8744} & \textbf{0.8658} & \textbf{0.0000} \\
    \bottomrule
  \end{tabular}
\end{table}

Tab.~\ref{tab:main_results} summarizes the five-domain continual trajectory under the oracle setting. Sequential FT suffers severe forgetting on both backbones. $\mathcal{F}$ reaches $22.05$~dB for Restormer and $15.83$~dB for NAFNet, and the final-stage average $A_B$ collapses to $13.62$~dB and $12.20$~dB respectively, far below the per-domain peaks captured in $\bar{A}$. This gap between $A_B$ and $\bar{A}$, exceeding $10$~dB on Restormer and $8$~dB on NAFNet, tracks the progressive overwriting of earlier-domain representations as each new degradation is absorbed. The standard continual-learning remedies fare little better under this degree of inter-degradation shift. EWC does not improve on naive fine-tuning, its $A_B$ of $12.00$~dB on Restormer and $11.63$~dB on NAFNet sitting at or below sequential FT, with forgetting reduced only on Restormer. LwF holds forgetting to $\mathcal{F}$ of $3.04$ and $0.22$~dB, but restricting a single shared model to every domain caps its final quality at $A_B$ of $17.59$ and $18.10$~dB. Regularization of this kind, whether over gradients or outputs, preserves competence in class-incremental classification but not when successive domains demand qualitatively different restoration behaviors.

RwF removes this collapse by construction. With zero forgetting, final $A_B$ rises to $28.87$~dB for Restormer, a $+15.25$~dB gain over sequential FT, and to $24.03$~dB for NAFNet ($+11.83$~dB), narrowing the gap between $A_B$ and $\bar{A}$ to under $1$~dB on both backbones, with SSIM following the same trend. These results are competitive with joint all-in-one training over the five domains, which reaches $26.72$ and $25.49$~dB while optimizing on all degradations at once. RwF-Restormer exceeds its joint counterpart and RwF-NAFNet trails it only slightly, so assigning each degradation an isolated path recovers the quality of joint optimization without revisiting past data, and on the stronger backbone surpasses it by sidestepping cross-degradation interference. The behavior is consistent across architectures, holding for both a Transformer and a convolutional backbone under identical hyperparameters, with Restormer leading NAFNet by close to $5$~dB under RwF.

\subsection{Domain-Agnostic Inference on Real Degradations}
\label{subsec:agnostic}

Under the domain-agnostic setting, the router must infer adapter identity (Sec.~\ref{subsec:routing}) from the degraded input alone. To test whether this transfers beyond the synthetic training corpus, we deploy RwF-Restormer on eleven canonical real-degradation benchmarks across all five domains without any retraining or recalibration: CBSD68~\cite{martin2001database}, Kodak24~\cite{franzen1999kodak}, and Urban100~\cite{huang2015single} with synthetic additive Gaussian noise at $\sigma{=}25$ for $\mathcal{D}_1$; RealBlur-J and RealBlur-R~\cite{rim2020real} for $\mathcal{D}_2$; Rain100H, Rain100L~\cite{yang2017deep}, and Test100~\cite{zhang2019image} for $\mathcal{D}_3$; SOTS-Indoor and SOTS-Outdoor~\cite{li2018benchmarking} for $\mathcal{D}_4$; and LOL-v1~\cite{wei2018deep} for $\mathcal{D}_5$, totaling $3,465$ images. Tab.~\ref{tab:domain_agnostic} reports per-benchmark routing accuracy and reconstruction quality.

\begin{table}[ht]
  \centering
  \small
  \caption{Domain-agnostic evaluation of RwF-Restormer on eleven real-degradation benchmarks. Routing accuracy and reconstruction quality are reported under both \emph{domain-agnostic} (Pred.) and \emph{oracle} settings.}
  \label{tab:domain_agnostic}
  \begin{tabular}{l l r c cc cc}
    \toprule
    & & & & \multicolumn{2}{c}{PSNR (dB) $\uparrow$}
    & \multicolumn{2}{c}{SSIM $\uparrow$} \\
    \cmidrule(lr){5-6} \cmidrule(lr){7-8}
    Domain & Test set & $N$ & ID Acc.\,(\%) & Pred. & Oracle & Pred. & Oracle \\
    \midrule
    $\mathcal{D}_1$ (noise)     & CBSD68          & 68    & 92.6  & 28.76 & 30.18 & 0.8448 & 0.8713 \\
                                & Kodak24         & 24    & 100.0 & 31.66 & 31.66 & 0.8754 & 0.8754 \\
                                & Urban100        & 100   & 90.0  & 28.74 & 30.75 & 0.8622 & 0.8906 \\
    \midrule
    $\mathcal{D}_2$ (blur)      & RealBlur-J      & 980   & 93.6  & 26.55 & 26.57 & 0.8349 & 0.8347 \\
                                & RealBlur-R      & 980   & 94.9  & 32.78 & 33.88 & 0.8962 & 0.9379 \\
    \midrule
    $\mathcal{D}_3$ (rain)      & Rain100H        & 100   & 100.0 & 16.92 & 16.92 & 0.4696 & 0.4696 \\
                                & Rain100L        & 100   & 83.0  & 30.18 & 32.13 & 0.8942 & 0.9291 \\
                                & Test100         & 98    & 70.4  & 21.20 & 21.64 & 0.6399 & 0.6762 \\
    \midrule
    $\mathcal{D}_4$ (haze)      & SOTS-Indoor     & 500   & 91.2  & 19.94 & 20.58 & 0.8728 & 0.8946 \\
                                & SOTS-Outdoor    & 500   & 72.4  & 25.50 & 27.93 & 0.9305 & 0.9625 \\
    \midrule
    $\mathcal{D}_5$ (low-light) & LOL-v1          & 15    & 46.7  & 12.13 & 16.65 & 0.4495 & 0.7356 \\
    \midrule
    \multicolumn{2}{l}{\textbf{Overall}}                            & \textbf{3,465} & \textbf{89.5} & \textbf{26.96} & \textbf{27.90} & \textbf{0.8568} & \textbf{0.8808} \\
    \bottomrule
  \end{tabular}
\end{table}

\begin{wrapfigure}{r}{0.42\textwidth}
  \vspace{-\intextsep}
  \centering
  \includegraphics[width=0.2\textwidth]{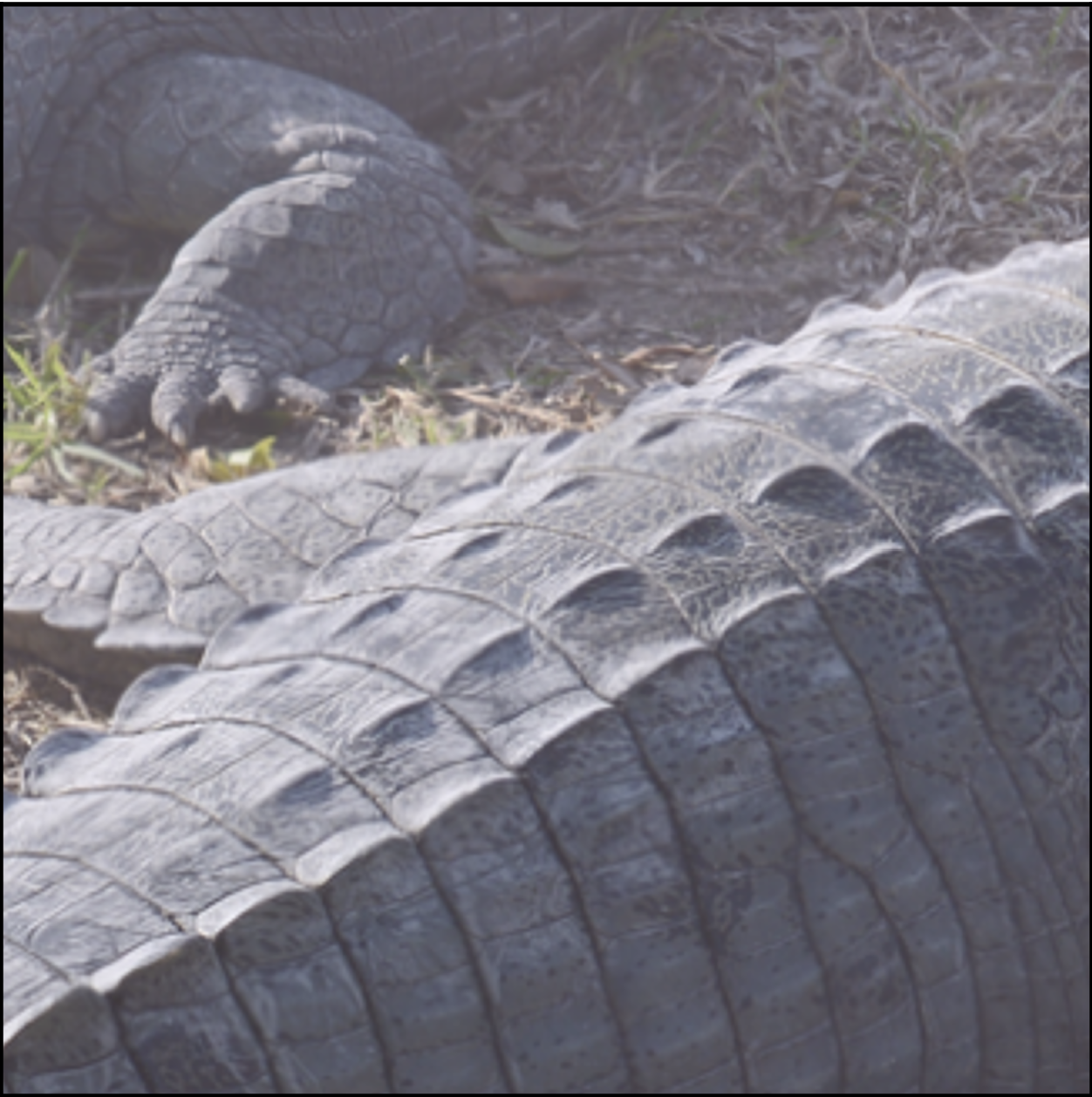}\hspace{1mm}%
  \includegraphics[width=0.2\textwidth]{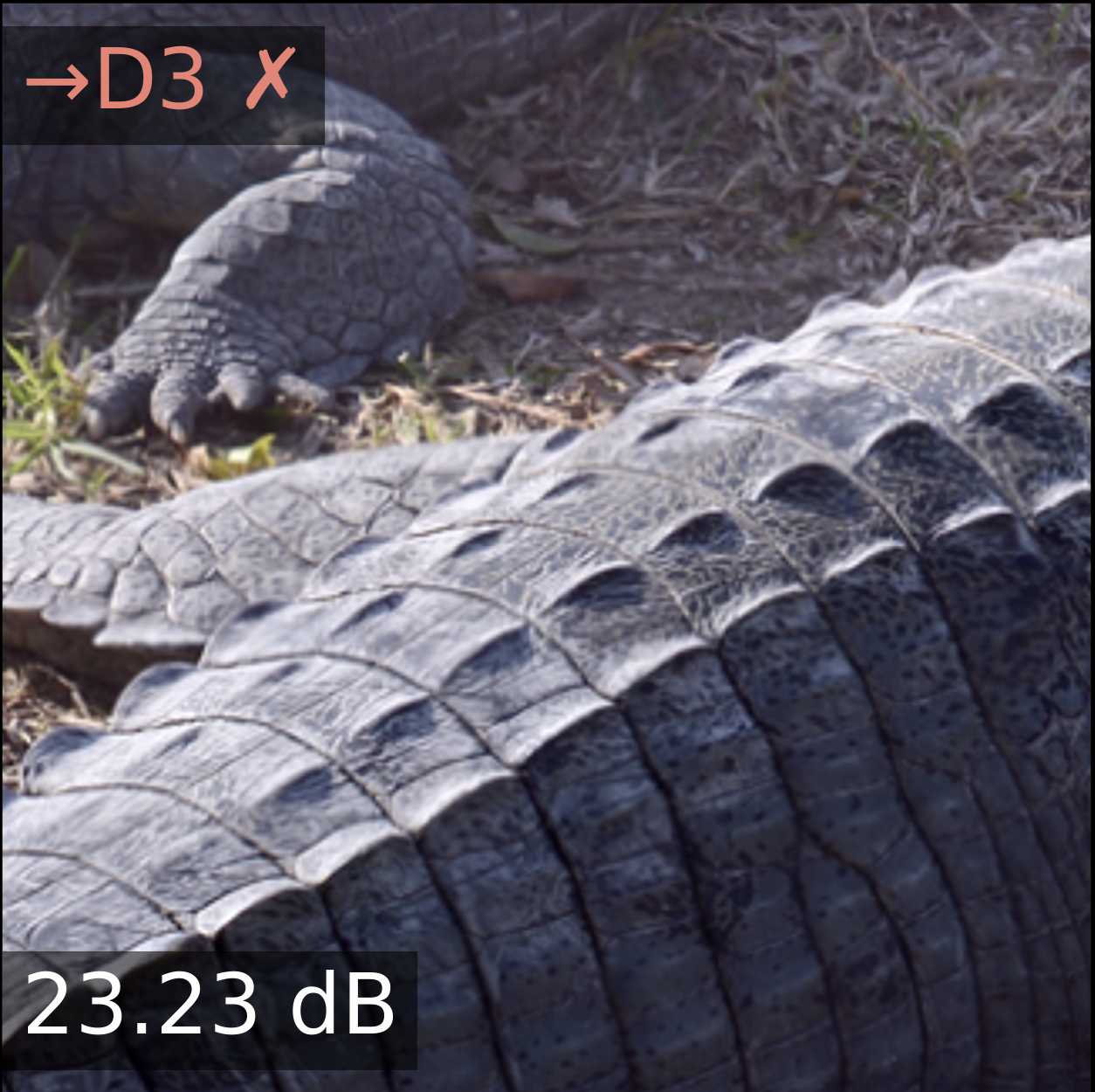}
  \caption{A $\mathcal{D}_4$ hazy input (left) misrouted to the $\mathcal{D}_3$ derain path is still largely restored (right).}
  \label{fig:misroute}
\end{wrapfigure}

Overall routing accuracy reaches $89.5\%$ 
(3,101 / 3,465 images correctly routed), with a residual $+0.94$~dB PSNR gap to the oracle ceiling. Seven of the eleven benchmarks exceed $90\%$ routing accuracy, and two (Kodak24 noise, Rain100H) achieve $100\%$ with predicted and oracle reconstruction quality matching exactly. The framework therefore transfers to real degradations without adaptation, supporting the content-controlled design rationale (Sec.~\ref{subsec:backbone}): isolating adapter learning from dataset-specific content priors lets the restoration paths generalize to unseen content of the same physical family. Crucially, misrouting is also graceful rather than catastrophic. As adjacent operators share restoration primitives (rain synthesis includes atmospheric veiling, low-light includes sensor noise; Sec.~\ref{subsec:synthesis}), a wrong route still applies a related correction rather than an arbitrary one, as in Fig.~\ref{fig:misroute}. 

Two cases, however, warrant attention. LOL-v1 is the weakest at $46.7\%$ routing accuracy, the regime our depth-discrimination argument predicts will be hard, since low-light's signature is a global intensity statistic that early encoder features partially equalize. Oracle routing on the same data recovers $16.65$~dB, so the failure is localized to routing, not the adapter. Rain100H shows the opposite imbalance, routing perfectly ($100\%$) but reaching only $16.92$~dB because the dataset's heavy-rain severity exceeds our $\mathcal{D}_3$ synthesis distribution, an adapter generalization gap rather than a routing failure.

\subsection{Ablation: Feature Readout Depth and Aggregation}
\label{subsec:ablation}

Two design choices govern the quality of the routing embedding: where along the backbone to read out the feature map, and how to summarize the resulting spatial tensor into a fixed-length vector. We ablate each independently on RwF-Restormer using the synthetic DIV2K-derived test set, which provides controlled comparison across configurations on identical content. The real-benchmark transfer of the selected configuration is reported in Sec.~\ref{subsec:agnostic}.

\begin{table}[ht]
\centering
\footnotesize
\begin{minipage}[t]{0.56\linewidth}
\centering
\caption{Routing quality vs.\ feature readout depth on RwF-Restormer.}
\label{tab:ablation_depth}
\begin{tabular}{l cc}
\toprule
Readout location  & ID Acc.\,(\%) & Oracle gap (dB) \\
\midrule
Encoder stage 1 (Ours)  & \textbf{79.7} & \textbf{+2.01} \\
Encoder stage 2  & 70.7 & +2.54 \\
Encoder stage 3  & 69.2 & +2.81 \\
Latent (bottleneck) & 58.8 & +4.14 \\
\bottomrule
\end{tabular}
\end{minipage}
\hfill
\begin{minipage}[t]{0.38\linewidth}
\centering
\caption{Routing quality vs.\ embedding aggregation.}
\label{tab:ablation_norm}
\setlength{\tabcolsep}{4pt}
\begin{tabular}{l c}
\toprule
Aggregation  & ID Acc.\,(\%) \\
\midrule
GAP only            & 73.9 \\
GAP + std (joint $\ell_2$)    & 78.4 \\
GAP + std (sep.\ $\ell_2$, Ours)     & \textbf{79.7} \\
\bottomrule
\end{tabular}
\end{minipage}
\end{table}

Tab.~\ref{tab:ablation_depth} reports routing accuracy and oracle PSNR gap as a function of readout depth, holding aggregation fixed at GAP + std with separate $\ell_2$ normalization. Routing quality degrades monotonically with depth: encoder stage 1 achieves $79.7\%$ accuracy ($+2.01$~dB gap), while the bottleneck collapses to $58.8\%$ and $+4.14$~dB. This reverses the classification-CL convention that deeper pre-classifier features yield more discriminative prototypes. A restoration backbone, by design, suppresses the very signal that distinguishes the domains, so late features grow informative about the clean image but less so about what was removed. Aggregating across scales, as adopted for degradation classification in DCPT~\cite{hu2025universal}, does not overcome this. Concatenating stages 1--3 yields $77.1\%$ accuracy and adding the latent stage drops it to $72.0\%$, both below the single shallow readout. This proves that combining scales only reintroduces the deeper features that dilute the degradation cue.

Tab.~\ref{tab:ablation_norm} reports the effect of aggregation at the fixed encoder-stage-1 readout. GAP alone reaches $73.9\%$ accuracy. Adding per-channel spatial standard deviation (GAP + std, jointly $\ell_2$-normalized) lifts it to $78.4\%$, confirming that second-order spatial statistics carry meaningful degradation signal beyond the mean, and normalizing the two moments separately before concatenation lifts it further to $79.7\%$. We adopt the separately-normalized variant on the basis of its higher domain-identification accuracy.

\subsection{Qualitative Analysis}
\label{subsec:qualitative}

\begin{figure}[ht]
    \centering
    \includegraphics[width=0.85\textwidth]{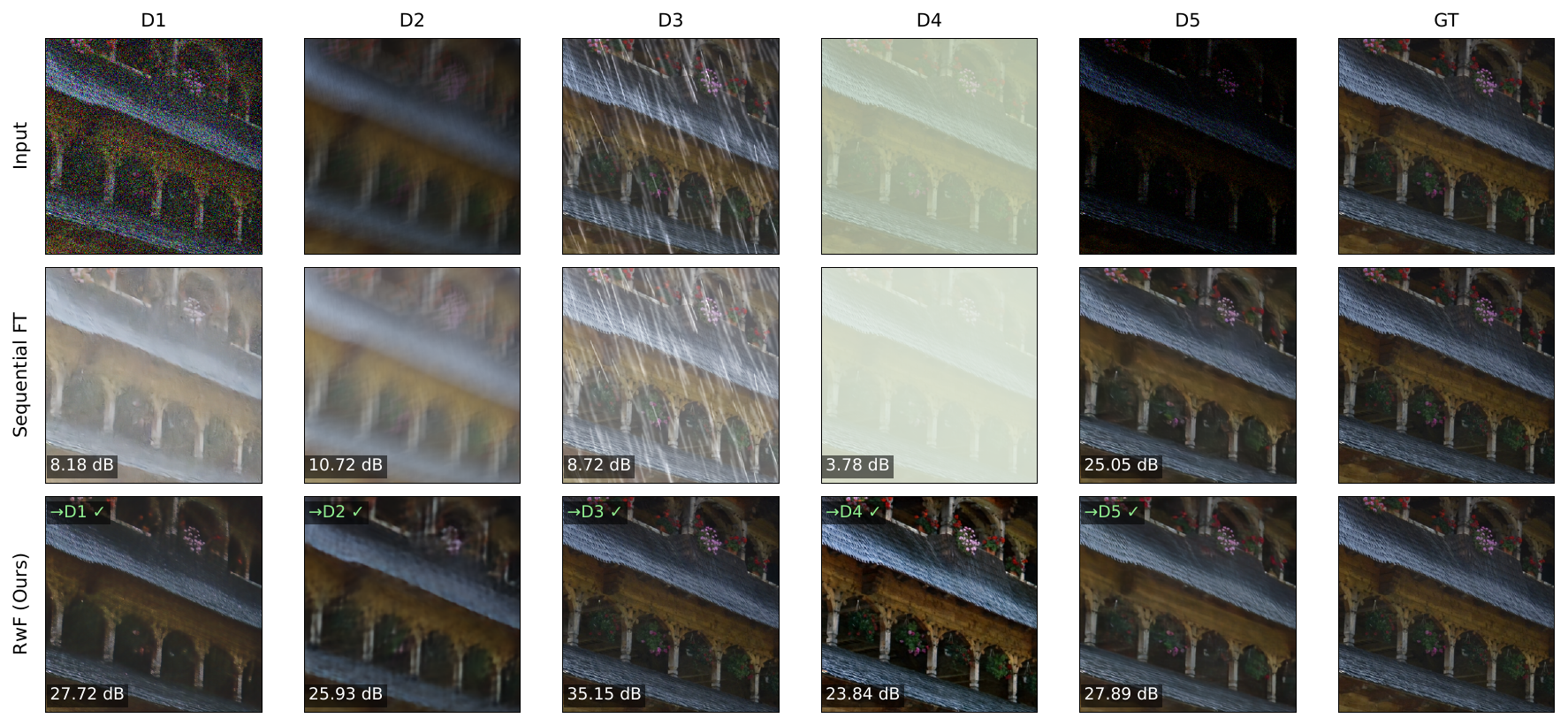}
    \caption{Qualitative comparison across all five degradation domains for a shared scene. \textbf{Top:} degraded inputs. \textbf{Middle:} sequentially fine-tuned Restormer after the full $\mathcal{D}_1{\rightarrow}\mathcal{D}_5$ schedule. \textbf{Bottom:} RwF-Restormer under domain-agnostic inference, with the router's prediction ($\rightarrow\mathcal{D}_t$\,\checkmark). Per-image PSNR is shown in the lower-left of each restored panel.}
    \label{fig:qualitative}
\end{figure}

Fig.~\ref{fig:qualitative} visualizes the practical consequence of the quantitative gap in Tab.~\ref{tab:main_results} on a single shared scene degraded under each of the five operators. Sequential fine-tuning recovers only $\mathcal{D}_5$, the most recently trained domain (25.05~dB). On every earlier domain, the final model outputs heavily artifacted reconstructions, collapsing to 3.78~dB on $\mathcal{D}_4$ where subsequent training stages have overwritten the model's haze-removal capability. RwF, by contrast, produces visually consistent reconstructions across all five domains (23.84 -- 35.15~dB), with the prototype router correctly selecting the appropriate restoration path for every input despite the substantial appearance differences between degraded variants.

\section{Conclusion}
\label{sec:conclusion}

We formulated multi-degradation image restoration as a continual domain-incremental learning problem and introduced Restoring without Forgetting (RwF), a parameter-efficient framework that freezes a denoising-pretrained backbone, attaches network-spanning low-rank adapter paths under strict parameter isolation, and routes unknown inputs to the matching path via unsupervised prototype matching. On a content-controlled five-domain benchmark spanning noise, blur, rain, haze, and low-light, RwF eliminates catastrophic forgetting by construction and improves final-stage average PSNR by up to $+15.25$~dB over naive sequential fine-tuning, on both a Transformer and a convolutional backbone. The framework transfers without modification to eleven canonical real-degradation benchmarks (3,465 images), reaching $89.5\%$ routing accuracy and a $+0.94$~dB oracle PSNR gap, establishing, to our knowledge, the first systematic baseline of its kind. Two directions remain open. First, routing accuracy degrades on degradations whose signature is primarily global rather than spatially localized, low-light enhancement in particular. A degradation-aware embedding head or a differentiable router could absorb the remaining accuracy gap on such regimes. Second, designing adapter pathway primitives that explicitly capture degradation-specific structure through frequency-domain projections, kernel-aware re-parameterizations, or richer cross-block interactions, is a promising direction for further per-domain quality gains, particularly on real benchmarks whose distribution exceeds the synthesis range of the training corpus.

\bibliography{ref}
\end{document}